\documentclass{article}

\usepackage{arxiv}

\usepackage[utf8]{inputenc} 
\usepackage[T1]{fontenc}    
\usepackage{lmodern}        
\usepackage{hyperref}       
\usepackage{url}            
\usepackage{booktabs}       
\usepackage{amsfonts}       
\usepackage{nicefrac}       
\usepackage{microtype}      
\usepackage{lipsum}
\usepackage{graphicx}

\title{Deep Reinforcement Learning for Conservation Decisions}

\author{
    Marcus Lapeyrolerie
   \\
    Department of Environmental Science, Policy, and Management \\
    University of California, Berkeley \\
  Berkeley, California \\
  \texttt{\href{mailto:mlapeyro@berkeley.edu}{\nolinkurl{mlapeyro@berkeley.edu}}} \\
   \And
    Melissa S. Chapman
   \\
    Department of Environmental Science, Policy, and Management \\
    University of California, Berkeley \\
  Berkeley, California \\
  \texttt{\href{mailto:mchapman@berkeley.edu}{\nolinkurl{mchapman@berkeley.edu}}} \\
   \And
    Kari E. A. Norman
   \\
    Department of Environmental Science, Policy, and Management \\
    University of California, Berkeley \\
  Berkeley, California \\
  \texttt{\href{mailto:kari.norman@berkeley.edu}{\nolinkurl{kari.norman@berkeley.edu}}} \\
   \And
    Carl Boettiger
   \\
    Department of Environmental Science, Policy, and Management \\
    University of California, Berkeley \\
  Berkeley, California \\
  \texttt{\href{mailto:cboettig@berkeley.edu}{\nolinkurl{cboettig@berkeley.edu}}
(corresponding author)} \\
  }

\newlength{\csllabelwidth}
\newlength{\cslhangindent}
  {}%
  {\par}
\newenvironment{CSLReferences}[3] 
 {
  \setlength{\parindent}{0pt}
  \ifodd #1 \everypar{\setlength{\hangindent}{\cslhangindent}}\ignorespaces\fi
  \ifnum #2 > 0
  \setlength{\parskip}{#2\baselineskip}
  \fi
 }%
 {}
\usepackage{calc} 

\usepackage{amsmath}
\usepackage{amssymb}

\begin{document}
\maketitle

\def\tightlist{}

\begin{abstract}
Can machine learning help us make better decisions about a changing
planet? In this paper, we illustrate and discuss the potential of a
promising corner of machine learning known as \emph{reinforcement
learning} (RL) to help tackle the most challenging conservation decision
problems. RL is uniquely well suited to conservation and global change
challenges for three reasons: (1) RL explicitly focuses on designing an
agent who \emph{interacts} with an environment which is dynamic and
uncertain, (2) RL approaches do not require massive amounts of data, (3)
RL approaches would utilize rather than replace existing models,
simulations, and the knowledge they contain. We provide a conceptual and
technical introduction to RL and its relevance to ecological and
conservation challenges, including examples of a problem in setting
fisheries quotas and in managing ecological tipping points. Four
appendices with annotated code provide a tangible introduction to
researchers looking to adopt, evaluate, or extend these approaches.
\end{abstract}


\hypertarget{introduction}{%
\section{Introduction}\label{introduction}}

Advances in both available data and computing power are opening the door
for machine learning (ML) to play a greater role in addressing some of
our planet's most pressing environmental problems. But will ML
approaches really help us tackle our most pressing environmental
problems? From the growing frequency and intensity of wildfire (Moritz
et al. 2014), to over-exploited fisheries (Worm et al. 2006) and
declining biodiversity (Dirzo et al. 2014), to emergent zoonotic
pandemics (Dobson et al. 2020), the diversity and scope of environmental
problems are unprecedented. Applications of ML in ecology have to-date
illustrated the promise of two methods: \emph{supervised learning} (M.
B. Joseph 2020) and \emph{unsupervised learning} (Valletta et al. 2017).
However, the fields of ecology and conservation science have so far
overlooked the third and possibly most promising approach in the ML
triad: \emph{reinforcement learning} (RL). Three features distinguish RL
from other ML methods in ways that are particularly well suited to
addressing issues of global ecological change:

\begin{enumerate}
\def\labelenumi{\arabic{enumi})}
\tightlist
\item
  RL is explicitly focused on the task of selecting actions in an
  uncertain and changing environment to maximize some objective,
\item
  RL does not require massive amounts of representative sampled
  historical data,
\item
  RL approaches easily integrate with existing ecological models and
  simulations, which may be our best guide to understanding and
  predicting future possibilities.
\end{enumerate}

Despite relevance to decision making under uncertainty that could make
RL uniquely well suited for ecological and conservation problems, it has
so far seen little application in these fields. To date, the problems
considered by RL research have largely been drawn from examples in
robotic movement and games like Go and Starcraft (OpenAI et al. 2019;
Silver et al. 2018; Vinyals et al. 2019). Complex environmental problems
share many similarities to these tasks and games: the need to plan many
moves ahead given a large number of possible outcomes, to account for
uncertainty and to respond with contingency to the unexpected. RL agents
typically develop strategies by interacting with simulators, a practice
that should not be unsettling to ecologists since learning from
simulators is common across ecology. Rich, processes-based simulations
such as the SORTIE model in forest management (Pacala et al. 1996),
Ecopath with Ecosim in fisheries management (Steenbeek et al. 2016), or
climate change policy models (Nordhaus 1992) are already used to explore
scenarios and inform ecosystem management. Decision-theoretic approaches
based on optimal control techniques can only find the best strategy in
the simplest of ecological models; the so called ``curse of
dimensionality'' makes problems with a large number of states or actions
intractable by conventional methods (Wilson et al. 2006; Marescot et al.
2013; Ferrer-Mestres et al. 2021). Neural-network-based RL techniques,
referred to as \emph{deep RL}, have proven particularly effective in
problems involving complex, high-dimensional spaces that have previously
proven intractable to classical methods.

In this paper, we draw on examples from fisheries management and
ecological tipping points to illustrate how deep RL techniques can
successfully discover optimal solutions to previously solved management
scenarios and discover highly effective solutions to unsolved problems.
We demonstrate that RL-based approaches are capable but by no means a
magic bullet: reasonable solutions require careful design of training
environments, choice of RL algorithms, tuning and evaluation, as well as
substantial computational power. Our examples are intentionally simple,
aiming to provide a clear template for understanding that could be
easily extended to cover more realistic conditions. We include an
extensive appendices with carefully annotated code which should allow
readers to both reproduce and extend this analysis.

\hypertarget{rl-overview}{%
\section{RL overview}\label{rl-overview}}

All applications of RL can be divided into two components: an
\emph{environment} and an \emph{agent}. The \emph{environment} is
typically a computer simulation, though it is possible to use the real
world as the RL environment (Ha et al. 2020). The \emph{agent}, which is
often a computer program, continuously interacts with the environment.
At each time step, the agent observes the current \emph{state} of the
environment then performs an \emph{action}. As a result of this action,
the environment transitions to a new state and transmits a numerical
\emph{reward} signal to the agent. The goal of the agent is to learn how
to maximize its expected cumulative reward. The agent learns how to
achieve this objective during a period called \emph{training}. In
training, the agent \emph{explores} the available actions. Once the
agent comes across a highly rewarding sequence of observations and
actions, the agent will reinforce this behavior so that it is more
likely for the agent to \emph{exploit} the same high reward trajectory
in the future. Throughout this process, the agent's behavior is codified
into what is called a \emph{policy}, which describes what action an
agent should take for a given observation.

\hypertarget{rl-environments}{%
\subsection{RL Environments}\label{rl-environments}}

An environment is a mathematical function, computer program, or real
world experience that takes an agent's proposed \emph{action} as input
and returns an observation of the environment's current \emph{state} and
an associated \emph{reward} as output. In contrast to classical
approaches (Marescot et al. 2013), there are few restrictions on what
comprises a state or action. States and actions may be continuous or
discrete, completely or partially observed, single or multidimensional.
The main focus of building an RL environment, however, is on the
environment's transition dynamics and reward function. The designer of
the environment can make the environment follow any transition and
reward function provided that both are functions of the current state
and action. This ability allows RL environments to model a broad range
of decision making problems. For example, we can set the transitions to
be deterministic or stochastic. We can also specify the reward function
to be \emph{sparse}, whereby a positive reward can only be received
after a long sequence of actions, e.g.~the end point in a maze. In other
environments, an agent may have to learn to forgo immediate rewards (or
even accept an initial negative reward) in order to maximize the net
discounted reward as we illustrate in examples here.

\begin{figure}
\includegraphics[width=\linewidth]{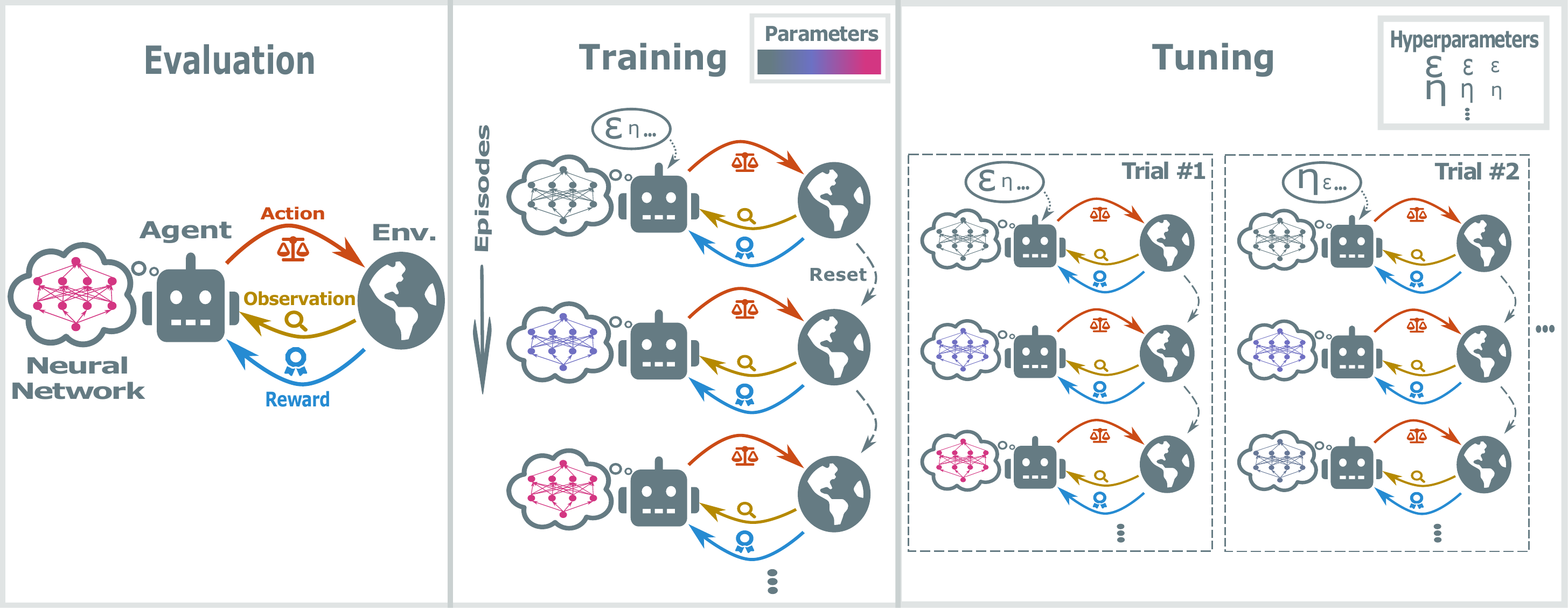}
\caption{Deep Reinforcement Learning:  A deep RL \emph{agent} uses a \emph{neural network} to select an \emph{action} in response to an \emph{observation} of the \emph{environment}, and receives a \emph{reward} from the environment as a result. During \emph{training}, the agent tries to maximize its cumulative reward by interacting with the environment and learning from experience. In the RL loop, the agent performs an action, then the environment returns a reward and an observation of the environment's state. The agent-environment loop continues until the environment reaches a terminal state, after which the environment will reset, causing a new \emph{episode} to begin. Across training episodes, the agent will continually update the \emph{parameters} in its neural network, so that the agent will select better actions. Before training starts, the researcher must input a set of \emph{hyperparameters} to the agent; hyperparameters direct the learning process and thus affect the outcome of training. A researcher finds the best set of hyperparameters during \emph{tuning}. Hyperparameter tuning consists of iterative \emph{trials}, in which the agent is trained with different sets of hyperparameters. At the end of a trial, the agent is evaluated to see which set of hyperparameters results in the highest cumulative reward. An agent is \emph{evaluated} by recording the cumulative reward over one episode, or the mean reward over multiple episodes. Within evaluation, the agent does not update its neural network; instead, the agent only uses a trained neural network to select actions.}
\end{figure}

The OpenAI \texttt{gym} software framework was created to address the
lack of standardization of RL environments and the need for better
benchmark environments to advance RL research (Brockman et al. 2016).
The \texttt{gym} framework defines a standard interface and methods by
which a developer can describe an arbitrary environment in a computer
program. This interface allows for the application of software agents
that can interact and learn in that environment without knowing anything
about the environment's internal details. Using the \texttt{gym}
framework, we turn existing ecological models into valid environmental
simulators that can be used with any RL agents. In Appendix C, we give
detailed instruction on how an OpenAI \texttt{gym} is constructed.

\hypertarget{deep-rl-agents}{%
\subsection{Deep RL Agents}\label{deep-rl-agents}}

To optimize the RL objective, agents either take a \emph{model-free} or
\emph{model-based} approach. The distinction is that \emph{model-free}
algorithms do not attempt to learn or use a model of the environment;
yet, \emph{model-based} algorithms employ a model of the environment to
achieve the RL objective. A trade-off between these approaches is that
when it is possible to quickly learn a model of the environment or the
model is already known, model-based algorithms tend to require much less
interaction with the environment to learn good-performing policies
(Janner et al. 2019; Sutton and Barto 2018). Yet, frequently, learning a
model of the environment is very difficult, and in these cases,
model-free algorithms tend to outperform (Janner et al. 2019).

Neural networks become useful in RL when the environment has a large
observation-action space\footnote{Conventionally, an observation-action
  space is considered to be large when it is non-tabular, i.e.~cannot be
  represented in a computationally tractable table.}, which happens
frequently with realistic decision-making problems. Whenever there is a
need for an agent to approximate some function, say a policy function,
neural networks can be used in this capacity due to their property of
being general function approximators (Hornik, Stinchcombe, and White
1989). Although there are other function approximators that can be used
in RL, e.g.~Gaussian processes (Grande, Walsh, and How 2014), neural
networks have excelled in this role because of their ability to learn
complex, non-linear, high dimensional functions and their ability to
adapt given new information (Arulkumaran et al. 2017). There is a
multitude of deep RL algorithms since there are many design choices that
can be made in constructing a deep RL agent -- see Appendix A for more
detail on these engineering decisions. In Table 1, we present some of
the more common deep RL algorithms which serve as good reference points
for the current state of deep RL.

\begin{table}
\begin{tabular}{l l l}
\hline
Abbreviation & Algorithm Name                                                    & Model \\
\hline
D-MPC        & Deep Model-Predictive Control (Lenz, Knepper, and Saxena 2015)    & Model-based \\
I2A          & Imagination-Augmented Agents (Weber et al. 2018)                  & Model-based \\
MBPO         & Model-based Policy Optimization (Janner et al. 2019)              & Model-based \\
DQN          & Deep Q Networks   (Mnih et al. 2015)                              & Model-free  \\
A2C          & Advantage Actor Critic  (Mnih et al. 2016)                        & Model-free  \\
A3C          & Asynchronous A2C  (Babaeizadeh et al. 2017)                       & Model-free  \\
TRPO         & Trust Region Policy Optimization  (Schulman, Levine,et al. 2017)  & Model-free  \\
PPO          & Proximal Policy Optimization  (Schulman, Wolski, et al. 2017)     & Model-free  \\
DDPG         & Deep Deterministic Policy Gradient   (Lillicrap et al. 2019)      & Model-free  \\
TD3          & Twin Delayed DDPG  (Fujimoto, Hoof, and Meger 2018)               & Model-free  \\
SAC          & Soft Actor Critic   (Haarnoja et al. 2018)                        & Model-free  \\
IMPALA       & Importance Weighted Actor Learner (Espeholt et al. 2018)          & Model-free  \\
\hline
\end{tabular}
\caption{Survey of common deep RL algorithms.}
\end{table}

Training a deep RL agent involves allowing the agent to interact with
the environment for potentially thousands to millions of time steps.
During training, the deep RL agent continually updates its neural
network parameters so that it will converge to an optimal policy. The
amount of time needed for an agent to learn high reward yielding
behavior cannot be predetermined and depends on a host of factors
including the complexity of the environment, the complexity of the
agent, and more. Yet, overall, it has been well established that deep RL
agents tend to be very sample inefficient (Gu et al. 2017), so it is
recommended to provide a generous training budget for these agents.

The deep RL agent controls the learning process with parameters called
\emph{hyperparameters}. Examples of hyperparameters include the step
size used for gradient ascent and the interval to interact with the
environment before updating the policy. In contrast, a weight or bias in
an agent's neural network is simply called a \emph{parameter}.
Parameters are learned by the agent, but the hyperparameters must be
specified by the RL practitioner. Since the optimal hyperparameters vary
across environments and can not be predetermined (Henderson et al.
2019), it is necessary to find a good-performing set of hyperparameters
in a process called hyperparameter tuning which uses standard
multi-dimensional optimization methods. We further discuss and show the
benefits of hyperparameter tuning in Appendix B.

\hypertarget{rl-objective}{%
\subsection{RL Objective}\label{rl-objective}}

The reinforcement learning environment is typically formalized as a
discrete-time partially observable Markov decision process (POMDP). A
POMDP is a tuple that consists of the following:

\begin{itemize}
\item
  \(\mathcal{S}\): a set of states called the state space
\item
  \(\mathcal{A}\): a set of actions called the action space
\item
  \(\Omega\) : a set of observations called the observation space
\item
  \(E(o_{t}| s_{t})\): an emission distribution, which accounts for an
  agent's observation being different from environment's state
\item
  \(T(s_{t+1}|s_t, a_t)\): a state transition operator which describes
  the dynamics of the system
\item
  \(r(s_t, a_t)\): a reward function
\item
  \(d_0(s_0)\): an initial state distribution
\item
  \(\gamma \in (0,1]\): a discount factor
\end{itemize}

The agent interacts with the environment in an iterative loop, whereby
the agent only has access to the observation space, action space and the
discounted reward signal, \(\gamma^t \, r(s_t, a_t)\). As the agent
interacts with the environment by selecting actions according to its
policy, \(\pi(a_t | o_t)\), the agent will create a trajectory,
\(\tau = (s_0, o_0, a_0, \dots, s_{H-1}, o_{H-1}, a_{H-1}, s_H)\). From
these definitions, we can provide an agent's trajectory distribution for
a given policy as,

\[
  p_\pi(\tau) = d_0(s_0) \prod_{t=0}^{H-1} \pi(a_t | o_t) \, E(o_{t}| s_{t}) \, T(s_{t+1} | s_t, a_t).
\]

The goal of reinforcement learning is for the agent to find an optimal
policy distribution, \(\pi^*(a_t|o_t)\), that maximizes the expected
return, \(J(\pi)\):

\[
  \pi^* = \underset{\pi}{\text{argmax}} \,\, \mathbb{E}_{\tau \sim p_\pi(\tau)}\Big[\sum_{t=0}^{H-1} \gamma^t r(s_t, a_t) \Big] =  \underset{\pi}{\text{argmax}} \,\, J (\pi).
\]

Although there are RL-based methods for infinite horizon problems,
i.e.~when \(H=\infty\), we will only present episodic or finite horizon
POMDPs in this study. In Appendix A, we will discuss in greater detail
how deep RL algorithms attempt to optimize the RL objective.

\hypertarget{examples}{%
\section{Examples}\label{examples}}

We provide two examples that illustrate the application and potential of
deep RL to ecological and conservation problems, highlighting both the
potential and the inherent challenges. Annotated code for these examples
may be found in Appendix B and at
\url{https://github.com/boettiger-lab/rl-intro}.

\hypertarget{sustainable-harvest}{%
\subsection{\texorpdfstring{Sustainable harvest
\label{sec:fishery}}{Sustainable harvest }}\label{sustainable-harvest}}

The first example focuses on the important but well-studied problem of
setting harvest quotas in fisheries management. This provides a natural
benchmark for deep RL approaches, since we can compare the RL solution
to the mathematical optimum directly. Determining fishing quotas is both
a critical ecological issue (Worm et al. 2006, 2009; Costello et al.
2016), and a textbook example that has long informed the management of
renewable resources within fisheries and beyond (Colin W. Clark 1990).

Given a population growth model that predicts the total biomass of a
fish stock in the following year as a function of the current biomass,
it is straightforward to determine what biomass corresponds to the
maximum growth rate of the stock, or \(B_{\textrm{MSY}}\), the biomass
at Maximum Sustainable Yield (MSY) (Schaefer 1954). When the population
growth rate is stochastic, the problem is slightly harder to solve, as
the harvest quota must constantly adjust to the ups and downs of
stochastic growth, but it is still possible to show the optimal strategy
merely seeks to maintain the stock at \(B_{\textrm{MSY}}\), adjusted for
any discounting of future yields (Reed 1979).

For illustrative purposes, we consider the simplest version of the
dynamic optimal harvest problem as outlined by (Colin W. Clark 1973)
(for the deterministic case) and (Reed 1979) (under stochastic
recruitment). The manager seeks to optimize the net present value
(discounted cumulative catch) of a fishery, observing the stock size
each year and setting an appropriate harvest quota in response. In the
classical approach, the best model of the fish population dynamics must
first be estimated from data, potentially with posterior distributions
over parameter estimates reflecting any uncertainty. From this model,
the optimal harvest policy -- that is, the function which returns the
optimal quota for each possible observed stock size -- can be determined
either by analytic (Reed 1979) or numerical (Marescot et al. 2013)
methods, depending on the complexity of the model. In contrast, a
model-free deep RL algorithm makes no assumption about the precise
functional form or parameter values underlying the dynamics -- it is in
principle agnostic to the details of the simulation.

We illustrate the deep RL approach using the model-free algorithm known
as Twin Delayed Deep Deterministic Policy Gradient or more simply, TD3
(Fujimoto, Hoof, and Meger 2018). A step-by-step walk-through for
training agents on this environment is provided in the Appendix. We
compare the resulting management, policy, and reward under the RL agent
to that achieved by the optimal management solution {[}Fig 2{]}. Despite
having no knowledge of the underlying model, the RL agent learns enough
to achieve nearly optimal performance.

\begin{figure}
\centering
\includegraphics{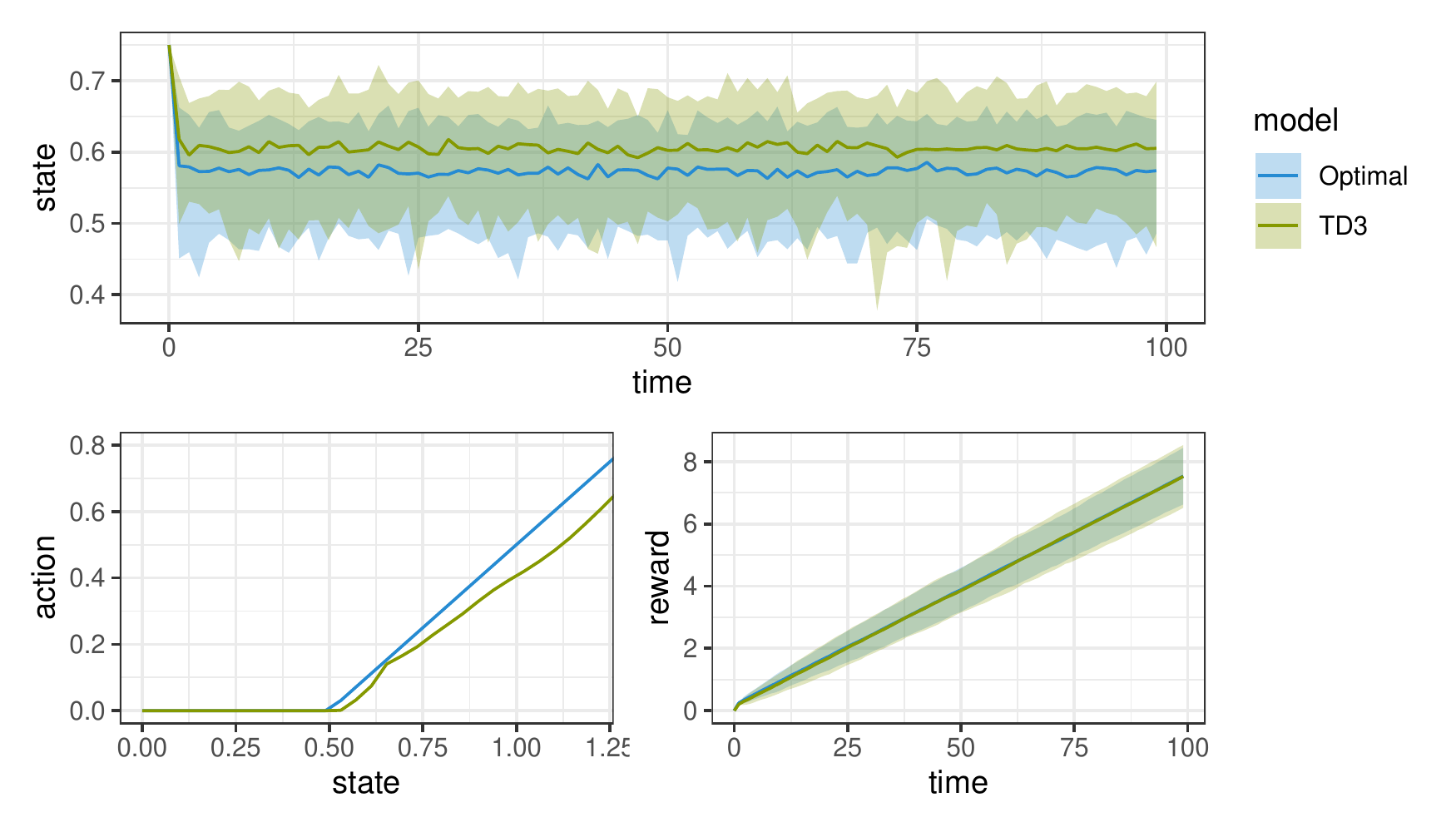}
\caption{Fisheries management using neural network agents trained with
RL algorithm TD3 compared to optimal management. Top panel: mean fish
population size over time across 100 replicates. Shaded region shows the
95\% confidence interval over simulations. Lower left: The optimal
solution is policy of constant escapement. Below the target escapement
of 0.5, no harvest occurs, while any stock above that level is
immediately harvested back down. The TD3 agent adopts a policy that
ceases any harvest below this level, while allowing a somewhat higher
escapement than optimal. TD3 achieves a nearly-optimal mean utility.}
\end{figure}

The cumulative reward (utility) realized across 100 stochastic
replicates is indistinguishable from that of the optimal policy {[}Fig
2{]}. Nevertheless, comparing the mean state over replicate simulations
reveals some differences in the RL strategy, wherein the stock is
maintained at a slightly higher-than-optimal biomass. Because our state
space and action space are sufficiently low-dimensional in this example,
we are also able to visualize the policy function directly, and compare
to the optimal policy {[}Fig 2{]}. This confirms that quotas tend to be
slightly lower than optimal, most notably at larger stock sizes. These
features highlight a common challenge in the design and training of RL
algorithms. RL cares only about improving the realized cumulative
reward, and may sometimes achieve this in unexpected ways. Because these
simulations rarely reach stock sizes at or above carrying capacity,
these larger stock sizes show a greater deviation from the optimal
policy than we observe at more frequently visited lower stock sizes.
Training these agents in a variety of alternative contexts can improve
there ability to generalize to other scenarios.

How would an RL agent be applied to empirical data? In principle, this
is straightforward. After we train an agent on a simulation environment
that approximates the fishery of interest, we can query the policy of
the agent to find a quota for the observed stock. To illustrate this, we
examine the quota that would be recommended by our newly trained RL
agent, above, against historical harvest levels of Argentine hake based
on stock assessments from 1986 - 2014 (RAM Legacy Stock Assessment
Database 2020, see Appendix D). Hake stocks showed a marked decline
throughout this period, while harvests decreased only in proportion
{[}Fig 3{]}. In contrast, our RL agent would have recommended
significantly lower quotas over most of the same interval, including the
closure of the fishery as stocks were sufficiently depleted. While we
have no way of knowing for sure if the RL quotas would have led to
recovery, let alone an optimal harvest rate, the contrast between those
recommended quotas and the historical catch is notable.

\begin{figure}
\centering
\includegraphics{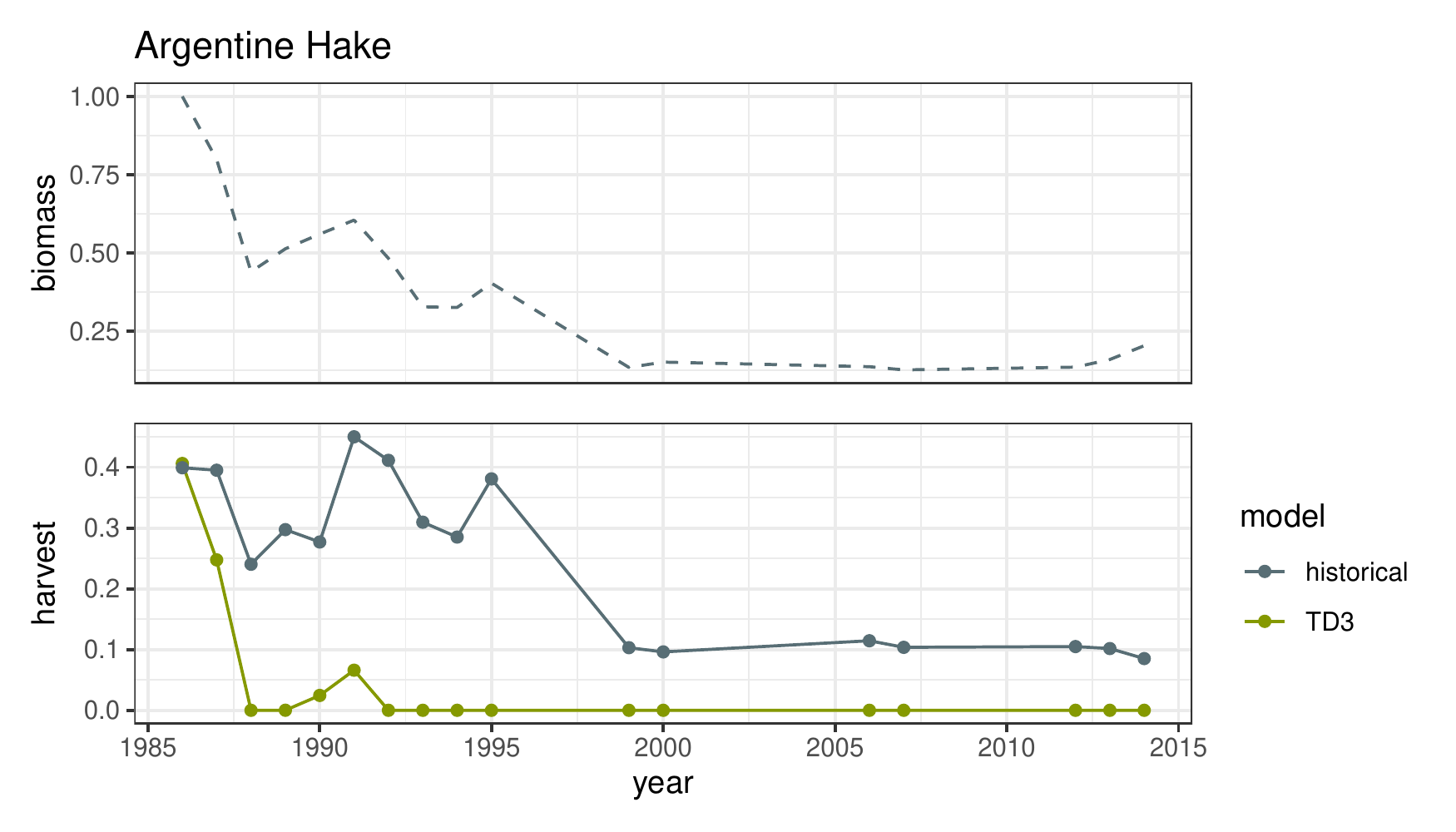}
\caption{Setting fisheries harvest quotas using Deep RL. Argentine Hake
fish stocks show a marked decline between 1986 and 2014 (upper panel).
Historical harvests (lower panel) declined only slowly in response to
consistently falling stocks, suggesting overfishing. In contrast,
RL-based quotas would have been set considerably lower than observed
harvests in each year of the data. As decline persists, the RL-based
management would have closed the fishery to future harvest until the
stock recovered.}
\end{figure}

This approach is not as different from conventional strategies as it may
seem. In a conventional approach, ecological models are first estimated
from empirical data, (stock assessments in the fisheries case). Quotas
can then be set based directly on these model estimates, or by comparing
alternative candidate ``harvest control rules'' (policies) against
model-based simulations of stock dynamics. This latter approach, known
in fisheries as Management Strategy Evaluation {[}MSE; Punt et al.
(2016){]} is already closely analogous to the RL process. Instead of
researchers evaluating a handful of control rules, the RL agent proposes
and evaluates a plethora of possible control rules autonomously.

\hypertarget{ecological-tipping-points}{%
\subsection{\texorpdfstring{Ecological Tipping Points
\label{sec:tipping}}{Ecological Tipping Points }}\label{ecological-tipping-points}}

Our second example focuses on a case for which we do not have an
existing, provably optimal policy to compare against. We consider the
generic problem of an ecosystem facing slowly deteriorating
environmental conditions which move the dynamics closer towards a
tipping point {[}Fig 4{]}. This model of a critical transition has been
posited widely in ecological systems, from the simple consumer-resource
model of (May 1977) on which our dynamics are based, to microbial
dynamics (Dai et al. 2012), lake ecosystem communities (Carpenter et al.
2011) to planetary ecosystems (Barnosky et al. 2012). On top of these
ecological dynamics we introduce an explicit ecosystem service model
quantifying the value of more desirable `high' state relative to the
`low' state. For simplicity, we assume a proportional benefit \(b\)
associated with the ecosystem state \(X(t)\). Thus when the ecosystem is
near the `high' equilibrium \(\hat X_H\), the corresponding ecosystem
benefit \(b \hat X_H\) is higher than at the low equilibrium, \(b x_L\),
consistent with the intuitive description of ecosystem tipping points
(Barnosky et al. 2012).

We also enumerate the possible actions which a manager may take in
response to environmental degradation. In the absence of any management
response, we assume the environment deteriorates at a fixed rate
\(\alpha\), which can be thought of as the incremental increase in
global mean temperature or similar anthropogenic forcing term.
Management can slow or even reverse this trend by choosing an opposing
action \(A_t\). We assume that large actions are proportionally more
costly than small actions, consistent with the expectation of
diminishing returns: taking the cost associated with an action \(A_t\)
as equal to \(c A_t^2\). Many alterations of these basic assumptions are
also possible: our \texttt{gym\_conservation} implements a range of
different scenarios with user-configurable settings. In each case, the
manager observes the current state of the system each year and must then
select the policy response that year.

\begin{figure}
\centering
\includegraphics{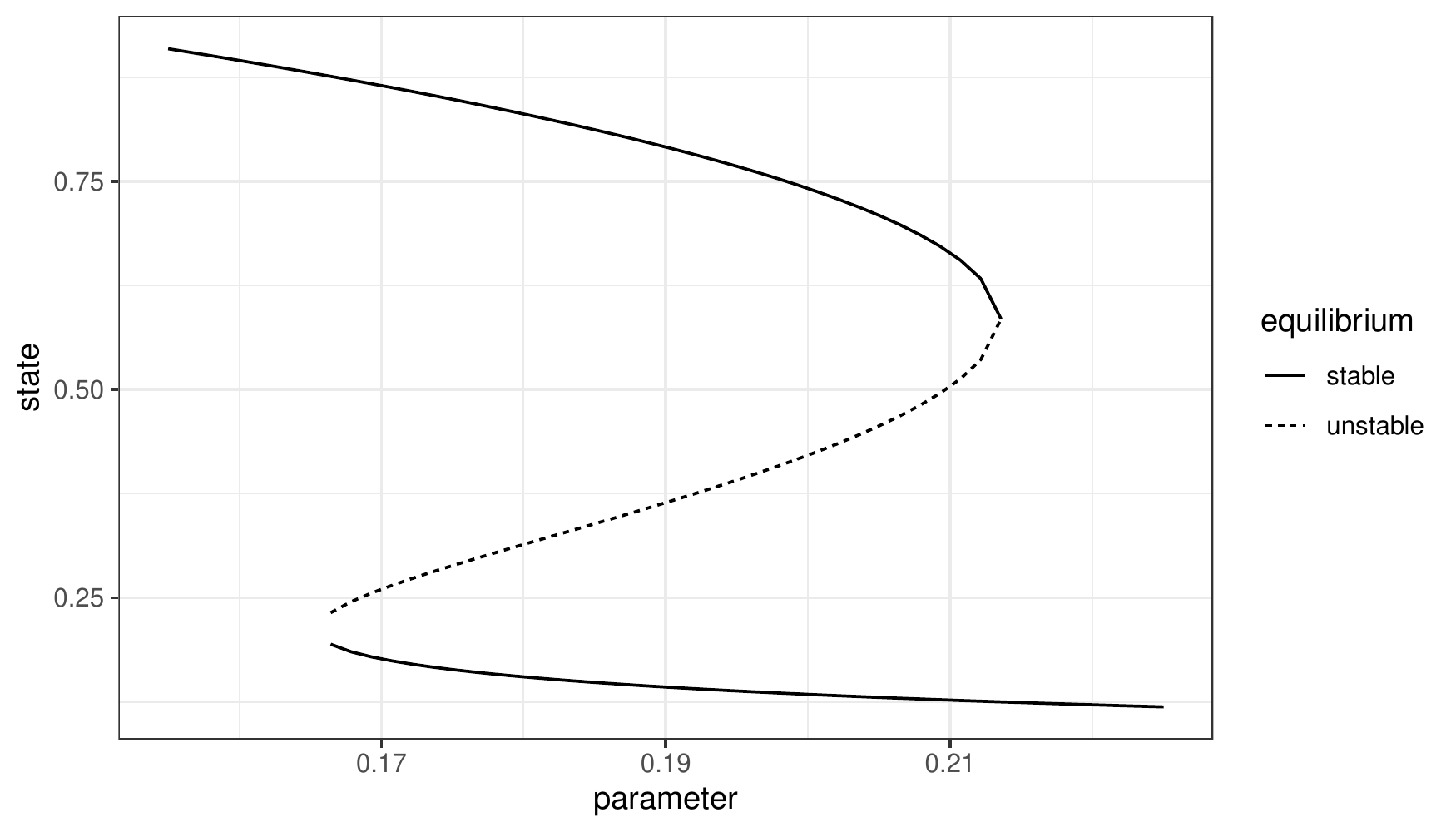}
\caption{Bifurcation diagram for tipping point scenario. The ecosystem
begins in the desirable `high' state under an evironmental parameter
(e.g.~global mean temperature, arbitrary units) of 0.19. In the absence
of conservation action, the environment worsens (e.g.~rising mean
temperature) as the parameter increases. This results in only a slow
degredation of the stable state, until the parameter crosses the tipping
point threshold at about 0.215, where the upper stable branch is
anihilated in a fold bifurcation and the system rapidly transitions to
lower stable branch, around state of 0.1. Recovery to the upper branch
requires a much greater conservation investment, reducing the parameter
all the way to 0.165 where the reverse bifurcation will carry it back to
the upper stable branch.}
\end{figure}

Because this problem involves a parameter whose value changes over time
(the slowly deteriorating environment), the resulting ecosystem dynamics
are not autonomous. This precludes our ability to solve for the optimal
management policy using classical theory such as for Markov Decision
Processes (MDP, (Marescot et al. 2013)), typically used to solve
sequential decision-making problems. However, it is often argued that
simple rules can achieve nearly optimal management of ecological
conservation objectives in many cases (Meir, Andelman, and Possingham
2004; Wilson et al. 2006; L. N. Joseph, Maloney, and Possingham 2009). A
common conservation strategy employs a fixed response level rather than
a dynamic policy which is toggled up or down each year: for example,
declaring certain regions as protected areas in perpetuity. An intuitive
strategy faced with an ecosystem tipping point would be `perfect
conservation,' in which the management response is perfectly calibrated
to counter-balance any further decline. While the precise rate of such
decline may not be known in practice (and will not be known to RL
algorithms before-hand either), it is easy to implement such a policy in
simulation for comparative purposes. We compare this rule-of-thumb to
the optimal policy found by training an agent using the TD3 algorithm.

\begin{figure}
\centering
\includegraphics{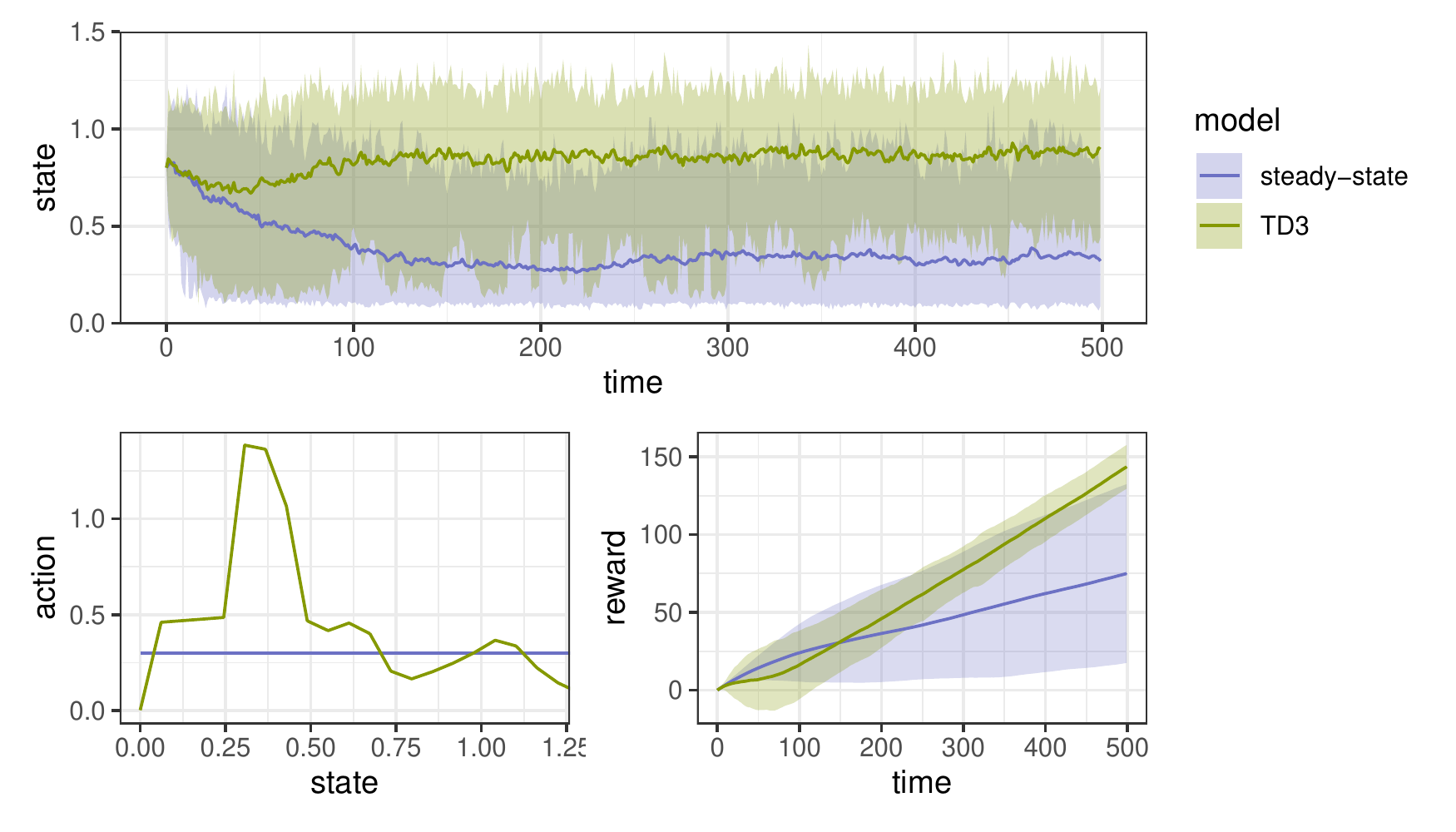}
\caption{Ecosystem dynamics under management using the steady-state
rule-of-thumb strategy compared to management using a neural network
trained using the TD3 RL algorithm. Top panel: mean and 95\% confidence
interval of ecosystem state over 100 replicate simulations. As more
replicates cross the tipping point threshold under steady-state
strategy, the mean slowly decreases, while the TD3 agent preserves most
replicates safely above the tipping point. Lower left: the policy
function learned using TD3 relative to the policy function under the
steady state. Lower right: mean rewards under TD3 management evenutally
exceed those expected under the steady-state strategy as a large initial
investment in conservation eventually pays off.}
\end{figure}

The TD3-trained agent proves far more successful in preventing chance
transitions across the tipping point, consistently achieving a higher
cumulative ecosystem service value across replicates than the
steady-state strategy.

Examining the replicate management trajectories and corresponding
rewards {[}Fig 5{]} reveal that the RL agent incurs significantly higher
costs in the initial phases of the simulation, dipping well below the
mean steady-state reward initially before exceeding it in the long run.
This initial investment then begins to pay off -- by about the 200th
time step the RL agent has surpassed the performance of the steady-state
strategy. The policy plot provides more intuition for the RL agent's
strategy: at very high state values, the RL agent opts for no
conservation action -- so far from the tipping point, no response is
required. Near the tipping point, the RL agent steeply ramps up the
conservation effort, and retains this effort even as the system falls
below the critical threshold, where a sufficiently aggressive response
can tip the system back into recovery. For a system at or very close to
the zero-state, the RL agent gives up, opting for no action. Recall that
the quadratic scaling of cost makes the rapid response of the TD3 agent
much more costly to achieve the same net environmental improvement
divided into smaller increments over a longer timeline. However, our RL
agent has discovered that the extra investment for a rapid response is
well justified as the risk of crossing a tipping point increases.

A close examination of the trajectories of individual simulations which
cross the tipping point under either management strategy {[}see Appendix
B{]} further highlights the difference between these two approaches.
Under the steady-state strategy, the system remains poised too close to
the tipping point: stochastic noise eventually drives most replicates
across the threshold, where the steady-state strategy is too weak to
bring them back once they collapse. As replicate after replicate
stochastically crashes, the mean state and mean reward bend increasingly
downwards. In contrast, the RL agent edges the system slightly farther
away from the tipping point, decreasing but not eliminating the chance
of a chance transition. In the few replicates that experience a critical
transition anyway, the RL agent usually responds with sufficient
commitment to ensure their recovery {[}Appendix B{]}. Only 3 out of 100
replicates degrade far enough for the RL agent to give up the high cost
of trying to rescue them. The RL agent's use of a more dynamic strategy
out-preforms the steady-state strategy. Numerous kinks visible in the RL
policy function also suggest that this solution is not yet optimal. Such
quirks are likely to be common features of RL solutions -- long as they
have minimal impact on realized rewards. Further tuning of
hyper-parameters, increased training, alterations or alternatives to the
training algorithm would likely be able to further improve upon this
performance.

\hypertarget{additional-environments}{%
\subsection{Additional Environments}\label{additional-environments}}

Ecology holds many open problems for deep RL. To extend the examples
presented here to reflect greater biological complexity or more
realistic decision scenarios, the transition, emission and/or reward
functions of the environment can be modified. We provide an initial
library of example environments at
\url{https://boettiger-lab.github.io/conservation-gym}. Some
environments in this library include a wildfire \texttt{gym} that poses
the problem of wildfire suppression with a cellular automata model, an
epidemic \texttt{gym} that examines timing of interventions to curb
disease spread, as well as more complex variations of the fishing and
conservation environments presented above.

\hypertarget{discussion}{%
\section{Discussion}\label{discussion}}

Ecological challenges facing the planet today are complex and outcomes
are both uncertain and consequential. Even our best models and best
research will never provide a crystal ball to the future, only better
elucidate possible scenarios. Consequently, that research must also
confront the challenge of making robust, resilient decisions in a
changing world. The science of ecological management and quantitative
decision-making has a long history (e.g. Schaefer 1954; Walters and
Hilborn 1978) and remains an active area of research (Wilson et al.
2006; Fischer et al. 2009; Polasky et al. 2011). However, the
limitations of classical methods such as optimal control frequently
constrain applications to relatively simplified models (Wilson et al.
2006), ignoring elements such as spatial or temporal heterogeneity and
autocorrelation, stochasticity, imperfect observations, age or state
structure or other sources of complexity that are both pervasive and
influential on ecological dynamics (Hastings and Gross 2012). Complexity
comes not only from the ecological processes but also the available
actions. Deep RL agents have proven remarkably effective in handling
such complexity, particularly when leveraging immense computing
resources increasingly available through advances in hardware and
software (Matthews 2018).

The rapidly expanding class of model-free RL algorithms is particularly
appealing given the ubiquitous presence of model uncertainty in
ecological dynamics. Rarely do we know the underlying functional forms
for ecological processes. Methods which must first assume something
about the structure or functional form of a process before estimating
the corresponding parameter can only ever be as good as those structural
assumptions. Frequently available ecological data is insufficient to
distinguish between possible alternative models (Knape and Valpine
2012), or the correct model may be non-identifiable with any amount of
data. Model-free RL approaches offer a powerful solution for this thorny
issue. Model-free algorithms have proven successful at learning
effective policies even when the underlying model is difficult or
impossible to learn (Pong et al. 2020), as long as simulations of
possible mechanisms are available.

The examples presented here only scrape the surface of possible RL
applications to conservation problems. The examples we have focused on
are intentionally quite simple, though it is worth remembering that
these very same simple models have a long history of relevance and
application in both research and policy contexts. Despite their
simplicity, the optimal strategy is not always obvious before hand,
however intuitive it may appear in retrospect. In the case of the
ecosystem tipping point scenario, the optimal strategy is unknown, and
the approximate solution found by our RL implementation could almost
certainly be improved upon. In these simple examples in which the
simulation implements a single model, training is analogous to classical
methods which take the model as given (Marescot et al. 2013). But
classical approaches can be difficult to generalize when the underlying
model is unknown. In contrast, the process of training an RL algorithm
on a more complex problem is no different than training on a simple one:
we only need access to a simulation which can generate plausible future
states in response to possible actions. This flexibility of RL could
allow us to attain better decision-making insight for our most realistic
ecological models such as the individual-based models used in the
management of forests and wildfire (Pacala et al. 1996; Moritz et al.
2014), disease (Dobson et al. 2020), marine ecosystems (Steenbeek et al.
2016), or global climate change (Nordhaus 1992).

Successfully applying RL to complex ecological problems is no easy task.
Even on relatively uncomplicated environments, training an RL agent can
be more challenging than expected due to an entanglement of reasons like
hyperparameter instability and poor exploration that can be very
difficult to resolve (Henderson et al. 2019; Berger-Tal et al. 2014). As
Section 5.1 and 5.2 illustrate, it is important to begin with simple
problems, including those for which an optimal strategy is already
known. Such examples provide important benchmarks to calibrate the
performance, tuning and training requirements of RL. Once RL agents have
mastered the basics, the examples can be easily extended into more
complex problems by changing the environment. Yet, even in the case that
an agent performs well on a realistic problem, there will be a range of
open questions in using deep RL to inform decision-making. Since deep
neural networks lack transparency (Castelvecchi 2016), can we be
confident that the deep RL agent will generalize its past experience to
new situations? Given that there have been many examples of reward
misspecification leading to undesirable behavior (Hadfield-Menell et al.
2020), what if we have selected an objective that unexpectedly causes
damaging behavior? A greater role of algorithms in conservation
decision-making also raises questions about ethics and power,
particularly when those algorithms are opaque or proprietary Chapman et
al. (2021).

Deep RL is still a very young field, where despite several landmark
successes, potential far outstrips practice. Recent advances in the
field have proven the potential of the approach to solve complex
problems (Silver et al. 2016, 2017, 2018; Mnih et al. 2015), but
typically leveraging large teams with decades of experience in ML and
millions of dollars worth of computing power (Silver et al. 2017).
Successes have so far been concentrated in applications to games and
robotics, not scientific and policy domains, though this is beginning to
change (Popova, Isayev, and Tropsha 2018; Zhou, Li, and Zare 2017).
Iterative improvements to well-posed public challenges have proven
immensely effective in the computer science community in tackling
difficult problems, which allow many teams with diverse expertise not
only to compete but to learn from each other (Villarroel, Taylor, and
Tucci 2013; Deng et al. 2009). By working to develop similarly
well-posed challenges as clear benchmarks, ecology and environmental
science researchers may be able to replicate that collaborative,
iterative success in cracking hard problems.

\pagebreak

\hypertarget{references}{%
\section*{References}\label{references}}
\addcontentsline{toc}{section}{References}

\hypertarget{refs}{}
\begin{CSLReferences}{1}{0}
\leavevmode\hypertarget{ref-briefsurveydrl}{}%
Arulkumaran, Kai, Marc Peter Deisenroth, Miles Brundage, and Anil
Anthony Bharath. 2017. {``A {Brief} {Survey} of {Deep} {Reinforcement}
{Learning}.''} \emph{IEEE Signal Processing Magazine} 34 (6): 26--38.
\url{https://doi.org/10.1109/MSP.2017.2743240}.

\leavevmode\hypertarget{ref-Barnosky2012}{}%
Barnosky, Anthony D., Elizabeth A. Hadly, Jordi Bascompte, Eric L.
Berlow, James H. Brown, Mikael Fortelius, Wayne M. Getz, et al. 2012.
{``Approaching a State Shift in {Earth}'s Biosphere.''} \emph{Nature}
486 (7401): 52--58. \url{https://doi.org/10.1038/nature11018}.

\leavevmode\hypertarget{ref-berger-tal_exploration-exploitation_2014}{}%
Berger-Tal, Oded, Jonathan Nathan, Ehud Meron, and David Saltz. 2014.
{``The {Exploration}-{Exploitation} {Dilemma}: {A} {Multidisciplinary}
{Framework}.''} \emph{PLOS ONE} 9 (4): e95693.
\url{https://doi.org/10.1371/journal.pone.0095693}.

\leavevmode\hypertarget{ref-brockman2016}{}%
Brockman, Greg, Vicki Cheung, Ludwig Pettersson, Jonas Schneider, John
Schulman, Jie Tang, and Wojciech Zaremba. 2016. {``{OpenAI} {Gym}.''}
\emph{arXiv:1606.01540 {[}Cs{]}}, June.
\url{http://arxiv.org/abs/1606.01540}.

\leavevmode\hypertarget{ref-Carpenter2011}{}%
Carpenter, Stephen R, J. J. Cole, Michael L Pace, Ryan D. Batt, William
A Brock, Timothy J. Cline, J. Coloso, et al. 2011. {``Early {Warnings}
of {Regime} {Shifts}: {A} {Whole}-{Ecosystem} {Experiment}.''}
\emph{Science (New York, N.Y.)} 1079 (April).
\url{https://doi.org/10.1126/science.1203672}.

\leavevmode\hypertarget{ref-castelvecchi_can_2016}{}%
Castelvecchi, Davide. 2016. {``Can We Open the Black Box of {AI}?''}
\emph{Nature News} 538 (7623): 20.
\url{https://doi.org/10.1038/538020a}.

\leavevmode\hypertarget{ref-chapman_promoting_2021}{}%
Chapman, Melissa, William Oestreich, Timothy H. Frawley, Carl Boettiger,
Sibyl Diver, Bianca Santos, Caleb Scoville, et al. 2021. {``Promoting
Equity in Scientific Recommendations for High Seas Governance.''}
Preprint. EcoEvoRxiv. \url{https://doi.org/10.32942/osf.io/jhbuz}.

\leavevmode\hypertarget{ref-Clark1973}{}%
Clark, Colin W. 1973. {``Profit Maximization and the Extinction of
Animal Species.''} \emph{Journal of Political Economy} 81 (4): 950--61.
\url{https://doi.org/10.1086/260090}.

\leavevmode\hypertarget{ref-Clark1990}{}%
Clark, Colin W. 1990. \emph{{Mathematical Bioeconomics: The Optimal
Management of Renewable Resources, 2nd Edition}}. Wiley-Interscience.

\leavevmode\hypertarget{ref-Costello2016}{}%
Costello, Christopher, Daniel Ovando, Tyler Clavelle, C. Kent Strauss,
Ray Hilborn, Michael C. Melnychuk, Trevor A Branch, et al. 2016.
{``{Global fishery prospects under contrasting management regimes}.''}
\emph{Proceedings of the National Academy of Sciences} 113 (18):
5125--29. \url{https://doi.org/10.1073/pnas.1520420113}.

\leavevmode\hypertarget{ref-Dai2012}{}%
Dai, Lei, Daan Vorselen, Kirill S Korolev, and J. Gore. 2012. {``Generic
{Indicators} for {Loss} of {Resilience} {Before} a {Tipping} {Point}
{Leading} to {Population} {Collapse}.''} \emph{Science (New York, N.Y.)}
336 (6085): 1175--77. \url{https://doi.org/10.1126/science.1219805}.

\leavevmode\hypertarget{ref-imagenet}{}%
Deng, Jia, Wei Dong, Richard Socher, Li-Jia Li, Kai Li, and Li Fei-Fei.
2009. {``{ImageNet}: {A} Large-Scale Hierarchical Image Database.''} In
\emph{2009 {IEEE} {Conference} on {Computer} {Vision} and {Pattern}
{Recognition}}, 248--55. Miami, FL: IEEE.
\url{https://doi.org/10.1109/CVPR.2009.5206848}.

\leavevmode\hypertarget{ref-biodiversity}{}%
Dirzo, Rodolfo, Hillary S Young, Mauro Galetti, Gerardo Ceballos, Nick
JB Isaac, and Ben Collen. 2014. {``Defaunation in the Anthropocene.''}
\emph{Science} 345 (6195): 401--6.

\leavevmode\hypertarget{ref-covid}{}%
Dobson, Andrew P., Stuart L. Pimm, Lee Hannah, Les Kaufman, Jorge A.
Ahumada, Amy W. Ando, Aaron Bernstein, et al. 2020. {``Ecology and
Economics for Pandemic Prevention.''} \emph{Science} 369 (6502):
379--81. \url{https://doi.org/10.1126/science.abc3189}.

\leavevmode\hypertarget{ref-impala}{}%
Espeholt, Lasse, Hubert Soyer, Remi Munos, Karen Simonyan, Volodymir
Mnih, Tom Ward, Yotam Doron, et al. 2018. {``{IMPALA}: {Scalable}
{Distributed} {Deep}-{RL} with {Importance} {Weighted} {Actor}-{Learner}
{Architectures}.''} \emph{arXiv:1802.01561 {[}Cs{]}}, June.
\url{http://arxiv.org/abs/1802.01561}.

\leavevmode\hypertarget{ref-ferrer-mestres_k-n-momdps_2021}{}%
Ferrer-Mestres, Jonathan, Thomas G. Dietterich, Olivier Buffet, and
Iadine Chades. 2021. {``K-{N}-{MOMDPs}: {Towards} {Interpretable}
{Solutions} for {Adaptive} {Management}.''} \emph{Proceedings of the
AAAI Conference on Artificial Intelligence} 35 (17): 14775--84.
\url{https://ojs.aaai.org/index.php/AAAI/article/view/17735}.

\leavevmode\hypertarget{ref-Fischer2009}{}%
Fischer, Joern, Garry D Peterson, Toby A. Gardner, Line J Gordon, Ioan
Fazey, Thomas Elmqvist, Adam Felton, Carl Folke, and Stephen Dovers.
2009. {``Integrating Resilience Thinking and Optimisation for
Conservation.''} \emph{Trends in Ecology \& Evolution} 24 (10): 549--54.
\url{https://doi.org/10.1016/j.tree.2009.03.020}.

\leavevmode\hypertarget{ref-TD3}{}%
Fujimoto, Scott, Herke van Hoof, and David Meger. 2018. {``Addressing
{Function} {Approximation} {Error} in {Actor}-{Critic} {Methods}.''}
\emph{arXiv:1802.09477 {[}Cs, Stat{]}}, October.
\url{http://arxiv.org/abs/1802.09477}.

\leavevmode\hypertarget{ref-grande14}{}%
Grande, Robert, Thomas Walsh, and Jonathan How. 2014. {``Sample
Efficient Reinforcement Learning with Gaussian Processes.''} In
\emph{Proceedings of the 31st International Conference on Machine
Learning}, edited by Eric P. Xing and Tony Jebara, 32:1332--40.
Proceedings of Machine Learning Research 2. Bejing, China: PMLR.
\url{http://proceedings.mlr.press/v32/grande14.html}.

\leavevmode\hypertarget{ref-gu_q-prop_2017}{}%
Gu, Shixiang, Timothy Lillicrap, Zoubin Ghahramani, Richard E. Turner,
and Sergey Levine. 2017. {``Q-{Prop}: {Sample}-{Efficient} {Policy}
{Gradient} with {An} {Off}-{Policy} {Critic}.''} \emph{arXiv:1611.02247
{[}Cs{]}}, February. \url{http://arxiv.org/abs/1611.02247}.

\leavevmode\hypertarget{ref-ha_learning_2020}{}%
Ha, Sehoon, Peng Xu, Zhenyu Tan, Sergey Levine, and Jie Tan. 2020.
{``Learning to {Walk} in the {Real} {World} with {Minimal} {Human}
{Effort}.''} \emph{arXiv:2002.08550 {[}Cs{]}}, November.
\url{http://arxiv.org/abs/2002.08550}.

\leavevmode\hypertarget{ref-hadfield-menell_inverse_2020}{}%
Hadfield-Menell, Dylan, Smitha Milli, Pieter Abbeel, Stuart Russell, and
Anca Dragan. 2020. {``Inverse {Reward} {Design}.''}
\emph{arXiv:1711.02827 {[}Cs{]}}, October.
\url{http://arxiv.org/abs/1711.02827}.

\leavevmode\hypertarget{ref-Hastings2012}{}%
Hastings, Alan, and Louis J. Gross, eds. 2012. \emph{Encyclopedia of
{Theoretical} {Ecology}}. Oakland, CA: University of California Press.

\leavevmode\hypertarget{ref-drlthatmatters}{}%
Henderson, Peter, Riashat Islam, Philip Bachman, Joelle Pineau, Doina
Precup, and David Meger. 2019. {``Deep {Reinforcement} {Learning} That
{Matters}.''} \emph{arXiv:1709.06560 {[}Cs, Stat{]}}, January.
\url{http://arxiv.org/abs/1709.06560}.

\leavevmode\hypertarget{ref-universalapproximators}{}%
Hornik, Kurt, Maxwell Stinchcombe, and Halbert White. 1989.
{``Multilayer Feedforward Networks Are Universal Approximators.''}
\emph{Neural Networks} 2 (5): 359--66.
\url{https://doi.org/10.1016/0893-6080(89)90020-8}.

\leavevmode\hypertarget{ref-mbpo}{}%
Janner, Michael, Justin Fu, Marvin Zhang, and Sergey Levine. 2019.
{``When to {Trust} {Your} {Model}: {Model}-{Based} {Policy}
{Optimization}.''} \emph{arXiv:1906.08253 {[}Cs, Stat{]}}, November.
\url{http://arxiv.org/abs/1906.08253}.

\leavevmode\hypertarget{ref-Possingham2009}{}%
Joseph, Liana N., Richard F. Maloney, and Hugh P. Possingham. 2009.
{``Optimal {Allocation} of {Resources} Among {Threatened} {Species}: A
{Project} {Prioritization} {Protocol}.''} \emph{Conservation Biology} 23
(2): 328--38. \url{https://doi.org/10.1111/j.1523-1739.2008.01124.x}.

\leavevmode\hypertarget{ref-joseph_neural_2020}{}%
Joseph, Maxwell B. 2020. {``Neural Hierarchical Models of Ecological
Populations.''} \emph{Ecology Letters} 23 (4): 734--47.
\url{https://doi.org/10.1111/ele.13462}.

\leavevmode\hypertarget{ref-knape_are_2012}{}%
Knape, Jonas, and Perry de Valpine. 2012. {``Are Patterns of Density
Dependence in the {Global} {Population} {Dynamics} {Database} Driven by
Uncertainty about Population Abundance?''} \emph{Ecology Letters} 15
(1): 17--23. \url{https://doi.org/10.1111/j.1461-0248.2011.01702.x}.

\leavevmode\hypertarget{ref-lenz_deepmpc_2015}{}%
Lenz, Ian, Ross Knepper, and Ashutosh Saxena. 2015. {``{DeepMPC}:
{Learning} {Deep} {Latent} {Features} for {Model} {Predictive}
{Control}.''} In \emph{Robotics: {Science} and {Systems} {XI}}.
Robotics: Science; Systems Foundation.
\url{https://doi.org/10.15607/RSS.2015.XI.012}.

\leavevmode\hypertarget{ref-ddpg}{}%
Lillicrap, Timothy P., Jonathan J. Hunt, Alexander Pritzel, Nicolas
Heess, Tom Erez, Yuval Tassa, David Silver, and Daan Wierstra. 2019.
{``Continuous Control with Deep Reinforcement Learning.''}
\emph{arXiv:1509.02971 {[}Cs, Stat{]}}, July.
\url{http://arxiv.org/abs/1509.02971}.

\leavevmode\hypertarget{ref-Marescot2013}{}%
Marescot, Lucile, Guillaume Chapron, Iadine Chadès, Paul L. Fackler,
Christophe Duchamp, Eric Marboutin, and Olivier Gimenez. 2013.
{``Complex Decisions Made Simple: A Primer on Stochastic Dynamic
Programming.''} \emph{Methods in Ecology and Evolution} 4 (9): 872--84.
\url{https://doi.org/10.1111/2041-210X.12082}.

\leavevmode\hypertarget{ref-gpu_computing}{}%
Matthews, David. 2018. {``Supercharge Your Data Wrangling with a
Graphics Card.''} \emph{Nature} 562 (7725): 151--52.
\url{https://doi.org/10.1038/d41586-018-06870-8}.

\leavevmode\hypertarget{ref-May1977}{}%
May, Robert M. 1977. {``Thresholds and Breakpoints in Ecosystems with a
Multiplicity of Stable States.''} \emph{Nature} 269 (5628): 471--77.
\url{https://doi.org/10.1038/269471a0}.

\leavevmode\hypertarget{ref-Possingham2004}{}%
Meir, Eli, Sandy Andelman, and Hugh P. Possingham. 2004. {``Does
Conservation Planning Matter in a Dynamic and Uncertain World?''}
\emph{Ecology Letters} 7 (8): 615--22.
\url{https://doi.org/10.1111/j.1461-0248.2004.00624.x}.

\leavevmode\hypertarget{ref-A2C}{}%
Mnih, Volodymyr, Adrià Puigdomènech Badia, Mehdi Mirza, Alex Graves,
Timothy P. Lillicrap, Tim Harley, David Silver, and Koray Kavukcuoglu.
2016. {``Asynchronous {Methods} for {Deep} {Reinforcement}
{Learning}.''} \emph{arXiv:1602.01783 {[}Cs{]}}, June.
\url{http://arxiv.org/abs/1602.01783}.

\leavevmode\hypertarget{ref-DQN}{}%
Mnih, Volodymyr, Koray Kavukcuoglu, David Silver, Andrei A. Rusu, Joel
Veness, Marc G. Bellemare, Alex Graves, et al. 2015. {``Human-Level
Control Through Deep Reinforcement Learning.''} \emph{Nature} 518
(7540): 529--33. \url{https://doi.org/10.1038/nature14236}.

\leavevmode\hypertarget{ref-wildfire}{}%
Moritz, Max A., Enric Batllori, Ross A. Bradstock, A. Malcolm Gill, John
Handmer, Paul F. Hessburg, Justin Leonard, et al. 2014. {``Learning to
Coexist with Wildfire.''} \emph{Nature} 515 (7525): 58--66.
\url{https://doi.org/10.1038/nature13946}.

\leavevmode\hypertarget{ref-dice}{}%
Nordhaus, W. D. 1992. {``An {Optimal} {Transition} {Path} for
{Controlling} {Greenhouse} {Gases}.''} \emph{Science} 258 (5086):
1315--19. \url{https://doi.org/10.1126/science.258.5086.1315}.

\leavevmode\hypertarget{ref-openai_learning_2019}{}%
OpenAI, Marcin Andrychowicz, Bowen Baker, Maciek Chociej, Rafal
Jozefowicz, Bob McGrew, Jakub Pachocki, et al. 2019. {``Learning
{Dexterous} {In}-{Hand} {Manipulation}.''} \emph{arXiv:1808.00177 {[}Cs,
Stat{]}}, January. \url{http://arxiv.org/abs/1808.00177}.

\leavevmode\hypertarget{ref-sortie}{}%
Pacala, Stephen W., Charles D. Canham, John Saponara, John A. Silander,
Richard K. Kobe, and Eric Ribbens. 1996. {``Forest {Models} {Defined} by
{Field} {Measurements}: {Estimation}, {Error} {Analysis} and
{Dynamics}.''} \emph{Ecological Monographs} 66 (1): 1--43.
\url{https://doi.org/10.2307/2963479}.

\leavevmode\hypertarget{ref-Polasky2011}{}%
Polasky, Stephen, Stephen R. Carpenter, Carl Folke, and Bonnie Keeler.
2011. {``{Decision-making under great uncertainty: environmental
management in an era of global change}.''} \emph{Trends in Ecology {\&}
Evolution} 26 (8): 398--404.
\url{https://doi.org/10.1016/j.tree.2011.04.007}.

\leavevmode\hypertarget{ref-Pong2020}{}%
Pong, Vitchyr, Shixiang Gu, Murtaza Dalal, and Sergey Levine. 2020.
{``Temporal {Difference} {Models}: {Model}-{Free} {Deep} {RL} for
{Model}-{Based} {Control}.''} \emph{arXiv:1802.09081 {[}Cs{]}},
February. \url{http://arxiv.org/abs/1802.09081}.

\leavevmode\hypertarget{ref-popova_deep_2018}{}%
Popova, Mariya, Olexandr Isayev, and Alexander Tropsha. 2018. {``Deep
Reinforcement Learning for de Novo Drug Design.''} \emph{Science
Advances} 4 (7): eaap7885. \url{https://doi.org/10.1126/sciadv.aap7885}.

\leavevmode\hypertarget{ref-punt_management_2016}{}%
Punt, André E, Doug S Butterworth, Carryn L de Moor, José A A De
Oliveira, and Malcolm Haddon. 2016. {``Management Strategy Evaluation:
Best Practices.''} \emph{Fish and Fisheries} 17 (2): 303--34.
\url{https://doi.org/10.1111/faf.12104}.

\leavevmode\hypertarget{ref-ramlegacy}{}%
RAM Legacy Stock Assessment Database. 2020. {``RAM Legacy Stock
Assessment Database V4.491.''}
\url{https://doi.org/10.5281/zenodo.3676088}.

\leavevmode\hypertarget{ref-Reed1979}{}%
Reed, William J. 1979. {``{Optimal escapement levels in stochastic and
deterministic harvesting models}.''} \emph{Journal of Environmental
Economics and Management} 6 (4): 350--63.
\url{https://doi.org/10.1016/0095-0696(79)90014-7}.

\leavevmode\hypertarget{ref-Schaefer1954}{}%
Schaefer, Milner B. 1954. {``{Some aspects of the dynamics of
populations important to the management of the commercial marine
fisheries}.''} \emph{Bulletin of the Inter-American Tropical Tuna
Commission} 1 (2): 27--56. \url{https://doi.org/10.1007/BF02464432}.

\leavevmode\hypertarget{ref-trpo}{}%
Schulman, John, Sergey Levine, Philipp Moritz, Michael I. Jordan, and
Pieter Abbeel. 2017. {``Trust {Region} {Policy} {Optimization}.''}
\emph{arXiv:1502.05477 {[}Cs{]}}, April.
\url{http://arxiv.org/abs/1502.05477}.

\leavevmode\hypertarget{ref-scoville_algorithmic_2021}{}%
Scoville, Caleb, Melissa Chapman, Razvan Amironesei, and Carl Boettiger.
2021. {``Algorithmic Conservation in a Changing Climate.''}
\emph{Current Opinion in Environmental Sustainability} 51 (August):
30--35. \url{https://doi.org/10.1016/j.cosust.2021.01.009}.

\leavevmode\hypertarget{ref-alphaGo2016}{}%
Silver, David, Aja Huang, Chris J. Maddison, Arthur Guez, Laurent Sifre,
George van den Driessche, Julian Schrittwieser, et al. 2016.
{``Mastering the Game of {Go} with Deep Neural Networks and Tree
Search.''} \emph{Nature} 529 (7587): 484--89.
\url{https://doi.org/10.1038/nature16961}.

\leavevmode\hypertarget{ref-alphazero}{}%
Silver, David, Thomas Hubert, Julian Schrittwieser, Ioannis Antonoglou,
Matthew Lai, Arthur Guez, Marc Lanctot, et al. 2018. {``A General
Reinforcement Learning Algorithm That Masters Chess, Shogi, and {Go}
Through Self-Play.''} \emph{Science} 362 (6419): 1140--44.
\url{https://doi.org/10.1126/science.aar6404}.

\leavevmode\hypertarget{ref-alphaGoZero}{}%
Silver, David, Julian Schrittwieser, Karen Simonyan, Ioannis Antonoglou,
Aja Huang, Arthur Guez, Thomas Hubert, et al. 2017. {``Mastering the
Game of {Go} Without Human Knowledge.''} \emph{Nature} 550 (7676):
354--59. \url{https://doi.org/10.1038/nature24270}.

\leavevmode\hypertarget{ref-ecopath}{}%
Steenbeek, Jeroen, Joe Buszowski, Villy Christensen, Ekin Akoglu, Kerim
Aydin, Nick Ellis, Dalai Felinto, et al. 2016. {``Ecopath with {Ecosim}
as a Model-Building Toolbox: {Source} Code Capabilities, Extensions, and
Variations.''} \emph{Ecological Modelling} 319 (January): 178--89.
\url{https://doi.org/10.1016/j.ecolmodel.2015.06.031}.

\leavevmode\hypertarget{ref-suttonbarto}{}%
Sutton, Richard S, and Andrew G Barto. 2018. \emph{Reinforcement
Learning: An Introduction}. MIT press.

\leavevmode\hypertarget{ref-valletta_applications_2017}{}%
Valletta, John Joseph, Colin Torney, Michael Kings, Alex Thornton, and
Joah Madden. 2017. {``Applications of Machine Learning in Animal
Behaviour Studies.''} \emph{Animal Behaviour} 124 (February): 203--20.
\url{https://doi.org/10.1016/j.anbehav.2016.12.005}.

\leavevmode\hypertarget{ref-netflix_prize}{}%
Villarroel, J. Andrei, John E. Taylor, and Christopher L. Tucci. 2013.
{``Innovation and Learning Performance Implications of Free Revealing
and Knowledge Brokering in Competing Communities: Insights from the
{Netflix} {Prize} Challenge.''} \emph{Computational and Mathematical
Organization Theory} 19 (1): 42--77.
\url{https://doi.org/10.1007/s10588-012-9137-7}.

\leavevmode\hypertarget{ref-alphastar}{}%
Vinyals, Oriol, Igor Babuschkin, Wojciech M. Czarnecki, Michaël Mathieu,
Andrew Dudzik, Junyoung Chung, David H. Choi, et al. 2019.
{``Grandmaster Level in {StarCraft} {II} Using Multi-Agent Reinforcement
Learning.''} \emph{Nature} 575 (7782): 350--54.
\url{https://doi.org/10.1038/s41586-019-1724-z}.

\leavevmode\hypertarget{ref-Walters1978}{}%
Walters, Carl J, and Ray Hilborn. 1978. {``Ecological {Optimization} and
{Adaptive} {Management}.''} \emph{Annual Review of Ecology and
Systematics} 9 (1): 157--88.
\url{https://doi.org/10.1146/annurev.es.09.110178.001105}.

\leavevmode\hypertarget{ref-weber_imagination-augmented_2018}{}%
Weber, Théophane, Sébastien Racanière, David P. Reichert, Lars Buesing,
Arthur Guez, Danilo Jimenez Rezende, Adria Puigdomènech Badia, et al.
2018. {``Imagination-{Augmented} {Agents} for {Deep} {Reinforcement}
{Learning}.''} \emph{arXiv:1707.06203 {[}Cs, Stat{]}}, February.
\url{http://arxiv.org/abs/1707.06203}.

\leavevmode\hypertarget{ref-Possingham2006}{}%
Wilson, Kerrie A., Marissa F. McBride, Michael Bode, and Hugh P.
Possingham. 2006. {``Prioritizing Global Conservation Efforts.''}
\emph{Nature} 440 (7082): 337--40.
\url{https://doi.org/10.1038/nature04366}.

\leavevmode\hypertarget{ref-Worm2006}{}%
Worm, Boris, Edward B Barbier, Nicola Beaumont, J Emmett Duffy, Carl
Folke, Benjamin S Halpern, Jeremy B C Jackson, et al. 2006. {``{Impacts
of biodiversity loss on ocean ecosystem services.}''} \emph{Science (New
York, N.Y.)} 314 (5800): 787--90.
\url{https://doi.org/10.1126/science.1132294}.

\leavevmode\hypertarget{ref-Worm2009}{}%
Worm, Boris, Ray Hilborn, Julia K Baum, Trevor A Branch, Jeremy S
Collie, Christopher Costello, Michael J Fogarty, et al. 2009.
{``{Rebuilding global fisheries.}''} \emph{Science (New York, N.Y.)} 325
(5940): 578--85. \url{https://doi.org/10.1126/science.1173146}.

\leavevmode\hypertarget{ref-zhou_optimizing_2017}{}%
Zhou, Zhenpeng, Xiaocheng Li, and Richard N. Zare. 2017. {``Optimizing
{Chemical} {Reactions} with {Deep} {Reinforcement} {Learning}.''}
\emph{ACS Central Science} 3 (12): 1337--44.
\url{https://doi.org/10.1021/acscentsci.7b00492}.

\end{CSLReferences}

\bibliographystyle{unsrt}
\bibliography{references.bib}

\end{document}